\documentclass[11pt,a4paper]{article}
\usepackage[hyperref]{emnlp-ijcnlp-2019}
\usepackage{times}
\usepackage{latexsym}

\usepackage{url}
\aclfinalcopy %

\usepackage{color}
\usepackage{multirow}
\usepackage{multicol}
\usepackage{commath}
\usepackage{amsmath,bm}
\usepackage{tabularx}
\usepackage{algpseudocode}
\usepackage{dsfont}
\usepackage{booktabs}
\usepackage{pbox}
\usepackage{graphicx}
\usepackage{enumitem}
\usepackage{framed}
\usepackage{array}
\usepackage[linesnumbered,ruled,vlined]{algorithm2e}
\usepackage[linesnumbered]{algorithm2e}
\usepackage[detect-all]{siunitx}
\usepackage{amssymb}
\usepackage{comment}
\usepackage{siunitx} 
\usepackage{amsmath}
\usepackage{color}                  %

\newcommand{\eg}{{e.g.}}
\newcommand{\ie}{{i.e.}}
\newcommand{\vs}{{vs.}}
\newcommand{\hhide}[1]{}

\definecolor{themered}{HTML}{FF8375}
\usepackage{xparse}
\NewDocumentCommand{\heng}{ mO{} }{\textcolor{red}{\textsuperscript{\textit{Heng}}\textsf{\textbf{\small[#1]}}}}
\title{Improving Question Answering with External Knowledge}

\newcommand\blfootnote[1]{%
  \begingroup
  \renewcommand\thefootnote{}\footnote{#1}%
  \addtocounter{footnote}{-1}%
  \endgroup
}

\newcommand{\RNum}[1]{\uppercase\expandafter{\romannumeral #1\relax}}

\author{
 Xiaoman Pan\textsuperscript{1}\textsuperscript{*} ~~ Kai Sun\textsuperscript{2}\textsuperscript{*} ~~ Dian Yu\textsuperscript{3} ~~Jianshu Chen\textsuperscript{3} \\
 ~~ \textbf{Heng Ji}\textsuperscript{1} ~~ \textbf{Claire Cardie}\textsuperscript{2} ~~ \textbf{Dong Yu}\textsuperscript{3}\\
 \textsuperscript{1}University of Illinois at Urbana-Champaign, Champaign, IL, USA  \\
 \textsuperscript{2}Cornell University, Ithaca, NY, USA  \\
 \textsuperscript{3}Tencent AI Lab, Bellevue, WA, USA\\
}

\date{}

\begin{document}
\maketitle

\blfootnote{* Equal contribution. This work was conducted when the two authors were at Tencent AI Lab, Bellevue, WA.}
                           
\begin{abstract}

We focus on multiple-choice question answering (QA) tasks in subject areas such as science, where we require both broad background knowledge and the facts from the given subject-area reference corpus. In this work, we explore simple yet effective methods for exploiting two sources of external knowledge for subject-area QA. The first enriches the original subject-area reference corpus with relevant text snippets extracted from an open-domain resource (\ie, Wikipedia) that cover potentially ambiguous concepts in the question and answer options. As in other QA research, the second method simply increases the amount of training data by appending additional in-domain subject-area instances.

Experiments on three challenging multiple-choice science QA tasks (\ie, ARC-Easy, ARC-Challenge, and OpenBookQA) demonstrate the effectiveness of our methods: in comparison to the previous state-of-the-art, we obtain absolute gains in accuracy of up to $8.1\%$, $13.0\%$, and $12.8\%$, respectively. While we observe consistent gains when we introduce knowledge from Wikipedia, we find that employing additional QA training instances is not uniformly helpful: performance degrades when the added instances exhibit a higher level of difficulty than the original training data. As one of the first studies on exploiting unstructured external knowledge for subject-area QA, we hope our methods, observations, and discussion of the exposed limitations may shed light on further developments in the area.

\end{abstract}

\section{Introduction}

To answer questions relevant to a given text (\eg, a document or a book), human readers often rely on a certain amount of broad background knowledge obtained from sources outside of the text~\cite{mcnamara2004istart,salmeron2006reading}. It is perhaps not surprising then, that machine readers also require knowledge external to the text itself to perform well on question answering (QA) tasks.   
We focus on multiple-choice QA tasks in subject areas such as science, in which facts from the given reference corpus (\eg, a textbook) need to be combined with broadly applicable external knowledge to select the correct answer from the available options~\cite{clark2016combining,clark2018think,mihaylov2018can}. For convenience, we call these \textbf{subject-area QA} tasks. 

\begin{table}[h!]
\centering
\small
\begin{tabular}{p{0.2cm}p{2.6cm}p{0.2cm}p{2.6cm}}
\toprule
\multicolumn{4}{l}{\textbf{Question}: a magnet will stick to \underline{~~~~}?}
 \\
\textbf{A}.   &  a belt buckle. $\checkmark$ & \textbf{B}. & a wooden table. \\
\textbf{C}.   &  a plastic cup. & \textbf{D}.   &  a paper plate.     \\
\bottomrule
\end{tabular}
\caption{A sample problem from a multiple-choice QA task OpenBookQA~\cite{mihaylov2018can} in a scientific domain ($\checkmark$: correct answer option).}
\label{tab:sample1}
\end{table}

To correctly answer the question in Table~\ref{tab:sample1}, for example, scientific facts\footnote{Ground truth facts are usually not provided in this kind of question answering tasks.} from the provided reference corpus --- \{\textit{``a magnet attracts magnetic metals through magnetism''} and \textit{``iron is always magnetic''}\}, as well as general world knowledge extracted from an external source such as \{\textit{``a belt buckle is often made of iron''} and \textit{``iron is metal''}\} are required.
Thus, these QA tasks provide suitable testbeds for evaluating external knowledge exploitation and intergration. %

Previous subject-area QA methods (\eg,~\cite{khot2017answering,zhang2018kg,zhong2018improving}) explore many ways of exploiting structured knowledge. Recently, we have seen that the framework of fine-tuning a pre-trained language model (\eg, GPT~\cite{radfordimproving} and BERT~\cite{bert2018}) outperforms previous state-of-the-art methods~\cite{mihaylov2018can,ni2018learning}. However, it is still not clear how to incorporate different sources of external knowledge, especially unstructured knowledge, into this powerful framework to further improve subject-area QA.

We investigate two sources of external knowledge (\ie, \textbf{open-domain} and \textbf{in-domain}), which have
proven effective for other types of QA tasks, by incorporating them into a pre-trained language model during the \textbf{fine-tuning} stage. First, we identify concepts in question and answer options and link these potentially ambiguous concepts to an \textbf{open-domain} resource that provides unstructured background information relevant to the concepts and used to enrich the original reference corpus (Section~\ref{sec:method:open-domain}). In comparison to previous work (\eg,~\cite{yadavalignment}), we perform information retrieval based on the enriched corpus instead of the original one to form a document for answering a question. Second, we increase the amount of training data by appending additional \textbf{in-domain} subject-area QA datasets (Section~\ref{sec:method:in-domain}).

We conduct experiments on three challenging multiple-choice science QA tasks where existing methods stubbornly continue to exhibit performance gaps in comparison with humans: ARC-Easy, ARC-Challenge~\cite{clark2016combining,clark2018think}, and OpenBookQA~\cite{mihaylov2018can}, which are collected from real-world science exams or carefully checked by experts. We fine-tune BERT~\cite{bert2018} in a two-step fashion (Section~\ref{sec:method:basic}). We treat entire Wikipedia as the \textbf{open-domain} external resource (Section~\ref{sec:method:open-domain}) and all the evaluated science QA datasets (question-answer pairs and reference corpora) except the target one as \textbf{in-domain} external resources (Section~\ref{sec:method:in-domain}). Experimental results show that we can obtain absolute gains in accuracy of up to $8.1\%$, $13.0\%$, and $12.8\%$, respectively, in comparison to the previous published state-of-the-art, demonstrating the effectiveness of our methods. We also analyze the gains and exposed limitations. While we observe consistent gains by introducing knowledge from Wikipedia, employing additional in-domain training data is not uniformly helpful: performance degrades when the added data exhibit a higher level of difficulty than the original training data (Section~\ref{sec:exp}).

\begin{figure*}[!h]
   \begin{center}
   \includegraphics[width=.95\linewidth]{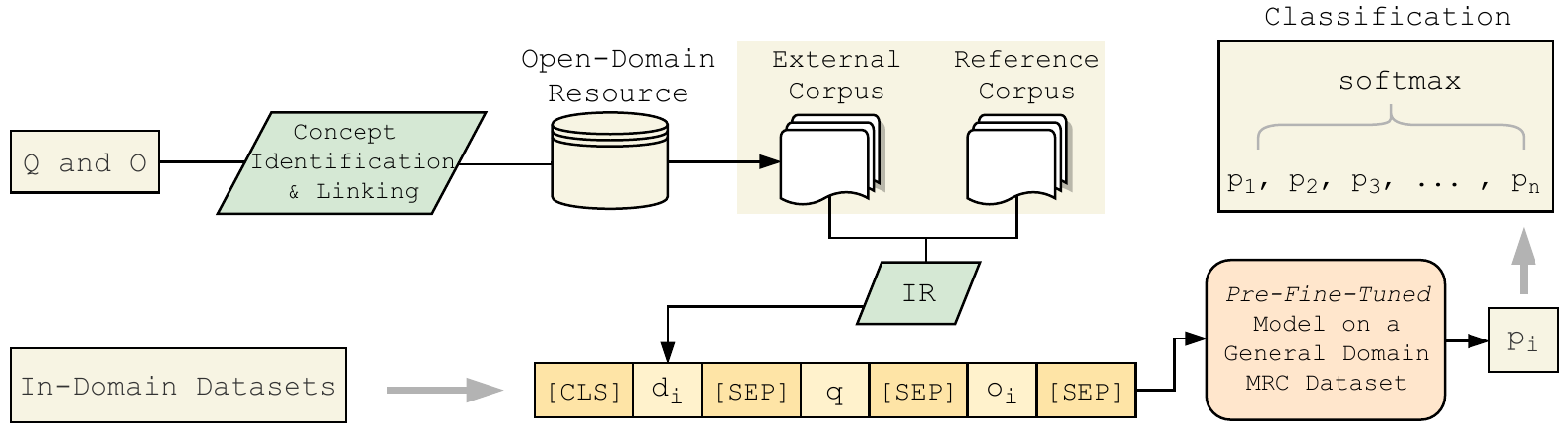}
   \end{center}
 \caption{Overview of our framework (IR: information retrieval; MRC: machine reading comprehension). $Q$, $O$, $q$, $o_i$, $d_i$, and $n$ denote the set of all questions, the set of all answer options, a question, one of the answer options associated with question $q$, the document (formed by retrieved sentences) associated with the ($q$, $o_i$) pair, and the number of answer options of $q$, respectively.}
 \label{fig:method:nnStruct}
\end{figure*}
To the best of our knowledge, this is the first work to incorporate external knowledge into a pre-trained model for improving subject-area QA. Besides, our promising results emphasize the importance of external unstructured knowledge for subject-area QA. We expect there is still much scope for further improvements by exploiting more sources of external knowledge, and we hope the present empirical study can serve as a new starting point for researchers to identify the remaining challenges in this area.

\section{Method}
\label{sec:method}
In this section, we first introduce our BERT-based QA baseline (Section~\ref{sec:method:basic}). Then, we present how we incorporate external open-domain (Section~\ref{sec:method:open-domain}) and in-domain (Section~\ref{sec:method:in-domain}) sources of knowledge into the baseline.

\subsection{Baseline Framework}
\label{sec:method:basic}

Given a question $q$, an answer option $o_i$, and a reference document $d_i$, we concatenate them with $@$ and $\#$ as the input sequence $@d_i\#q\#o_i\#$ to BERT~\cite{bert2018}, where $@$ and $\#$ stand for the classifier token $\texttt{[CLS]}$ and sentence separator token $\texttt{[SEP]}$ in BERT, respectively. A segmentation $\texttt{A}$ embedding is added to every token before $q$ (exclusive) and a segmentation $\texttt{B}$ embedding to every other token, where $\texttt{A}$ and $\texttt{B}$ are learned during the language model pretraining of BERT. For each instance in the ARC (Easy and Challenge) and OpenBookQA tasks, we use Lucene~\cite{lucene} to retrieve up to top $K$ sentences using the non-stop words in $q$ and $o_i$ as the query and then concatenate the retrieved sentences to form $d_i$~\cite{sun2018improving}. The final prediction for each question is obtained by a linear plus softmax layer over the output of the final hidden state of the first token in each input sequence. 

By default, we employ the following \textbf{two-step} fine-tuning approach unless explicitly specified. Following previous work~\cite{sun2018improving} based on GPT~\cite{radfordimproving}, we first fine-tune BERT~\cite{bert2018} on a large-scale multiple-choice machine reading comprehension dataset RACE~\cite{lai2017race} collected from English-as-a-foreign-language exams, which provides a ground truth reference document instead of a reference corpus for each question. Then, we further fine-tune the model on the target multiple-choice science QA datasets. For convenience, we call the model obtained after the first fine-tuning phase as a \textbf{pre-fine-tuned model}.

\subsection{Utilization of External Knowledge from an Open-Domain Resource}
\label{sec:method:open-domain}

Just as human readers activate their background knowledge related to the text materials~\cite{kendeou2007effects}, we link concepts identified in questions and answer options to an open-domain resource (\ie, Wikipedia) and provide machine readers with unstructured background information relevant to these concepts, used to enrich the original reference corpus. %

\begin{table}[t!]
\centering
\small
\begin{tabular}{p{.95\linewidth}}
\toprule
\textbf{Question}: Mercury, the planet nearest to the Sun, has extreme surface temperatures, ranging from $465^\circ$C in sunlight to $-180^\circ$C in darkness. Why is there such a large range of temperatures on Mercury? \\\\
\textbf{A}. The planet is too small to hold heat.                                                                                                                                                              \\
\textbf{B}. The planet is heated on only one side.                                                                                                                                                             \\
\textbf{C}. The planet reflects heat from its dark side.                                                                                                                                                       \\
\textbf{D}. The planet lacks an atmosphere to hold heat. $\checkmark$                        \\                           
\bottomrule                   
\end{tabular}
\caption{A sample problem from the ARC-Challenge dataset~\cite{clark2018think} ($\checkmark$: correct answer option).}
\label{tab:sample2}
\end{table}

\noindent \textbf{Concept Identification and Linking}: 
\label{sec:method:open-domain:cl}
We first extract concept mentions from texts.
Most mention extraction systems (\eg,~\citet{corenlp}) are trained using pre-defined classes in general domain such as \textsc{Person}, \textsc{Location}, and \textsc{Organization}.
However, in ARC and OpenBookQA, the vast majority of mentions are from scientific domains (\eg, \emph{``rotation''}, \emph{``revolution''}, \emph{``magnet''}, and \emph{``iron''}). Therefore, we simply consider all noun phrases as candidate concept mentions, which are extracted by a noun phrase chunker. For example, in the sample problem in Table~\ref{tab:sample2}, we extract concept mentions such as \emph{``Mercury''}.

Then each concept mention is disambiguated and linked to its corresponding concept (page) in Wikipedia.
For example, the ambiguous concept mention \emph{``Mercury''} in Table~\ref{tab:sample2} should be linked to the concept \texttt{Mercury\_(planet)} rather than \texttt{Mercury\_(element)} in Wikipedia. For concept disambiguation and linking, we simply adopt an existing unsupervised approach~\cite{Pan2015} that first selects high quality sets of concept \emph{collaborators} to feed a simple similarity measure (\ie, Jaccard) to link concept mentions.

\noindent \

\noindent \textbf{Reference Corpus Enrichment}:
\label{sec:method:open-domain:enrich}
We apply concept identification and linking to the text of all questions and answer options. Then, for each linked concept, we extract Wikipedia sentences that contain this concept and all sentences from the Wikipedia article of this concept without removing redundant information. For example, the following sentence in the Wikipedia article of \texttt{Mercury\_(planet)} is extracted: \emph{``Having almost no \textbf{atmosphere} to retain \textbf{heat}, it has surface temperatures that vary diurnally more than on any other planet in the Solar System.''}, which can serve as a reliable piece of evidence to infer the correct answer option D for the question in Table~\ref{tab:sample2}.

Most previous methods~\cite{khashabi2017learning,musa2018answering,ni2018learning,yadavalignment} perform information retrieval on the reference corpus to retrieve relevant sentences to form reference documents. In contrast, we retrieve relevant sentences from the \textbf{combination} of an open-domain resource and the original reference corpus to generate a reference document for each (question, answer option) pair. We still keep \textbf{up to top} $K$ sentences for each reference document (Section~\ref{sec:method:basic}). See the framework overview in Figure~\ref{fig:method:nnStruct}.

\subsection{Utilization of External Knowledge from In-Domain Data}
\label{sec:method:in-domain}

Since there are a relatively small number of training instances available for a single subject-area QA task (see Table~\ref{tab:eval:datasets}), instead of fine-tuning a pre-fine-tuned model on a single target dataset, we also investigate into fine-tuning a pre-fine-tuned model on multiple in-domain datasets simultaneously. For example, when we train a model for ARC-Challenge, we use the training set of ARC-Challenge together with the training, development, and test sets of ARC-Easy and OpenBookQA. We also explore two settings with and without merging the reference corpora from different tasks. We introduce more details and discussions in Section~\ref{sec:exp:settings} and Section~\ref{sec:exp:md}.

\section{Experiments and Discussions}
\label{sec:exp}

\subsection{Datasets}

In our experiment, we use RACE~\cite{lai2017race} --- the largest existing multiple-choice machine reading comprehension dataset collected from real and practical \textbf{language} exams --- in the pre-fine-tuning stage. Questions in RACE focus on evaluating linguistic knowledge acquisition of participants and are commonly used in previous methods~\cite{wang2018yuanfudao,sun2018improving}.

We evaluate the performance of our methods on three multiple-choice \textbf{science} QA datasets: ARC-Easy, ARC-Challenge, and OpenBookQA. ARC-Challenge and ARC-easy originate from the same set of exam problems collected from multiple sources. ARC-Challenge contains questions answered incorrectly by both a retrieval-based method and a word co-occurrence method, and the remaining questions form ARC-Easy. Questions in OpenBookQA are crowdsourced by turkers and then carefully filtered and modified by experts. See the statistics of these datasets in Table~\ref{tab:eval:datasets}. Note that for OpenBookQA, we do not utilize the accompanying auxiliary reference knowledge bases to ensure a fair comparison with previous work.

\begin{table}[t!]
\centering
\small
\begin{tabular}{lcccc}
\toprule
\bf Dataset       & \bf Train & \bf Dev  & \bf Test & \bf Total  \\
\midrule
RACE              & 87,866 & 4,887 & 4,934 & 97,687  \\
\midrule
ARC-Easy          & 2,251  & 570  & 2,376 & 5,197   \\
ARC-Challenge     & 1,119  & 299  & 1,172 & 2,590   \\
OpenBookQA        & 4,957  & 500  & 500  & 5,957   \\
\bottomrule
\end{tabular}
\caption{The number of questions in RACE and the multiple-choice subject-area QA datasets for evaluation: ARC-Easy, ARC-Challenge, and OpenBookQA.}%
\label{tab:eval:datasets}
\end{table}

\subsection{Experimental Settings}
\label{sec:exp:settings}

For the two-step fine-tuning framework, we use the uncased $\text{BERT}_\text{LARGE}$ released by~\newcite{bert2018} as the pre-trained language model. We set the batch size to $24$, learning rate to $2\times 10^{-5}$, and the maximal sequence length to $512$. When the input sequence length exceeds $512$, we truncate the longest sequence among $q$, $o_i$, and $d_i$ (defined in Section~\ref{sec:method:basic}).  We first fine-tune $\text{BERT}_\text{LARGE}$ for five epochs on RACE to get the pre-fine-tuned model and then further fine-tune the model for eight epochs on the target QA datasets in scientific domains. We show the accuracy of the pre-fine-tuned model on RACE in Table~\ref{tab:eval:bert_race}.
\begin{table}[t!]
\centering
\small
\begin{tabular}{lcc}
\toprule
\bf Dataset          & \bf Dev    & \bf Test \\
\midrule
RACE-M               & 76.7               & 76.6 \\
RACE-H               & 71.0               & 70.1 \\
RACE-M + RACE-H                & 72.7               & 72.0 \\
\bottomrule
\end{tabular}
\caption{Accuracy (\%) of the pre-fine-tuned model on the RACE dataset, which contains two subsets: RACE-M and RACE-H, representing problems collected from \textbf{m}iddle and \textbf{h}igh school language exams, respectively.}
\label{tab:eval:bert_race}
\end{table}

We use the noun phrase chunker in spaCy\footnote{\url{https://spacy.io/}.} to extract concept mentions. For information retrieval, we use the version 7.4.0 of Lucene~\cite{lucene} and set the maximum number of the retrieved sentences $K$ to $50$. We use the stop word list from NLTK~\cite{bird2004nltk}.

In addition, we design two slightly different settings for information retrieval. In \textbf{setting 1}, the original reference corpus of each dataset is independent. Formally, for each dataset $x\in D$, we perform information retrieval based on the corresponding original reference corpus of $x$ and/or the external corpus generated based on problems in $x$, where $D=\{\text{ARC-Easy}, \text{ARC-Challenge}, \text{OpenBookQA}\}$. In \textbf{setting 2}, all original reference corpora are integrated to further leverage external in-domain knowledge. Formally, for each dataset $x\in D$, we conduct information retrieval based on the given reference corpus of $D$ and/or the external corpus generated based on problems in $D$ instead of $x$.\footnote{\url{https://github.com/nlpdata/external}.}.

\subsection{Baselines}
\label{sec:baseline}

\begin{table*}[!t]
\centering
\small
\begin{tabular}{lcccl}

\toprule
\bf Method                                       & \bf ARC-E & \bf ARC-C & \bf OBQA  \\
\midrule
IR~\cite{clark2018think}                         & 62.6   & 20.3         & --     \\  
Odd-One-Out~\cite{mihaylov2018can}               & --     & --           & 50.2     \\
DGEM~\cite{khot2018scitail}                      & 59.0   & 27.1         & 24.4    \\   
KG$^2$~\cite{zhang2018kg}                        & --     & 31.7         & --       \\  %
AIR~\cite{yadav2018sanity}                       & 58.4   & 26.6         & --          \\ %
NCRF$++$~\cite{musa2018answering}                & 52.2   & 33.2         & --       \\ %
TriAN$++$~\cite{zhong2018improving}              & --     & 33.4         & --        \\  %
Two Stage Inference~\cite{pirtoaca2018improving} & 61.1   & 26.9         & --         \\  %
ET-RR~\cite{ni2018learning}                      &  --    & 36.6         & --       \\ 
GPT$^{\text{\RNum{2}}}$~\cite{radfordimproving,sun2018improving}     & 57.0   & 38.2       & 52.0    \\  
RS$^{\text{\RNum{2}}}$~\cite{sun2018improving}                     & 66.6   & 40.7         & 55.2      \\
\midrule

\textbf{Our BERT-Based Implementations}        &          &             &   \\ 
\textbf{Setting 1}       &          &             &    \\ 
Reference Corpus (RC) (\ie, BERT$^{\text{\RNum{2}}}$)       & 71.9     & 44.1       & 64.8 \\ 
External Corpus (EC)                           & 65.0     & 39.4        & 62.2 \\
RC + EC                                        & 73.3     & 45.0        & 65.2   \\
\textbf{Setting 2}        &          &             &        \\ 
Integrated Reference Corpus (IRC)              & 73.2     & 44.8        & 65.0      \\ 
Integrated External Corpus (IEC)               & 68.9     & 40.1        & 63.0   \\
IRC + IEC                                      & \bf 74.7 & 46.1        & 67.0   \\
IRC + MD                                       & 69.4     & 50.7        & 67.4   \\
IRC + IEC + MD                                 & 72.3     & \bf 53.7    & \bf 68.0    \\
\textbf{Human Performance}                &  --        &  --           &  91.7      \\

\bottomrule
\end{tabular}
\caption{Accuracy (\%) on the test sets of ARC-Easy, ARC-Challenge, and OpenBookQA datasets. RACE is used in the pre-fine-tuning stage for all the tasks (Section~\ref{sec:method:basic}). MD stands for fine-tuning on \textbf{m}ultiple target \textbf{d}atasets simultaneously (Section~\ref{sec:method:in-domain}). All results are single-model performance. GPT$^{\text{\RNum{2}}}$, RS$^{\text{\RNum{2}}}$, and BERT$^{\text{\RNum{2}}}$ are baselines that use two-step fine-tuning (Section~\ref{sec:baseline}). ARC-E: ARC-Easy; ARC-C: ARC-Challenge; OBQA: OpenBookQA.}
\label{tab:eval:transfer}
\end{table*}

Here we only briefly introduce three baselines (\ie, GPT$^{\text{\RNum{2}}}$, RS$^{\text{\RNum{2}}}$, and BERT$^{\text{\RNum{2}}}$) that all fine-tune a pre-trained language model on downstream tasks without substantial modifications to model architectures, which achieve remarkable success on many question answering tasks. Following the two-step fine-tuning framework (Section~\ref{sec:method:basic}), \textbf{all} three strong baselines use RACE in the first fine-tuning stage for a fair comparison. We will discuss the impacts of pre-fine-tuning on baseline model performance in Section~\ref{sec:exp:race}, noting that pre-fine-tuning is not the contribution of this work.

\noindent \textbf{GPT$^{\text{\RNum{2}}}$}: This baseline is based on fine-tuning a generative pre-trained transformer (GPT) language model~\cite{radfordimproving} instead of BERT~\cite{bert2018}.

\noindent \textbf{RS$^{\text{\RNum{2}}}$}: Based on GPT, general reading strategies (RS)~\cite{sun2018improving} are applied during the fine-tuning stage such as adding a trainable embedding into the text embedding of tokens relevant to the question and candidate answer options.

\noindent \textbf{BERT$^{\text{\RNum{2}}}$}: Based on BERT, this baseline is an exact implementation described in Section~\ref{sec:method:basic}.

\subsection{Main Results}
\label{sec:main_results}
We see consistent improvements in accuracy across all tasks after we enrich the reference corpus with relevant texts from Wikipedia to form new reference documents (\ie, $\text{RC} + \text{EC}$ and $\text{IRC} + \text{IEC}$ in Table~\ref{tab:eval:transfer}). Moreover, using only the extracted external corpus to perform information retrieval for reference document generation can achieve reasonable performance compared to using the original reference corpus, especially on the OpenBookQA dataset ($62.2\%$ \vs~$64.8\%$ under setting 1 and $63.0\%$ \vs~$65.0\%$ under setting 2). This indicates that we can extract reliable and relevant texts from external open-domain resources such as Wikipedia via linked concepts mentioned in Section~\ref{sec:method:open-domain}. Moreover, using the integrated corpus (\ie, setting $2$) consistently boosts the performance. Since the performance in setting $2$ (integrated corpus) is better than that in setting $1$ (independent corpus) based on our experiments, we take \textbf{setting 2} by default for discussions unless explicitly specified.

\begin{table*}[!t]
\centering
\small
\begin{tabular}{p{0.28\linewidth}p{0.35\linewidth}p{0.28\linewidth}}
\toprule
\bf Question & \bf Answer Options & \bf Sentence(s) From Wikipedia \\
\midrule
\multirow{4}{1.0\linewidth}{What boils at the boiling point?} 
& A. \textbf{\textit{Kool-Aid}}. $\checkmark$ &  \multirow{4}{1.0\linewidth}{\textbf{\textit{Kool-Aid}} is known as Nebraska's official soft drink. Common types of drinks include plain drinking \textbf{\textit{water}}, milk, coffee, tea, hot chocolate, juice and \textbf{\textit{soft drinks}}.} \\
&  B. Cotton. &  \\
&  C. Paper Towel. &  \\
&  D. Hair. &  \\
\\
\midrule
\multirow{4}{1.0\linewidth}{\textbf{\textit{Forest fires}} occur in many areas due to \textbf{\textit{drought conditions}}. If the drought conditions continue for a long period of time, which might cause the repopulation of trees to be threatened?} 
& A. a decrease in the \textbf{\textit{thickness of soil}}. $\checkmark$ &  \multirow{4}{1.0\linewidth}{It is highly resistant to \textbf{\textit{drought conditions}}, and provides excellent fodder; and has also been used in controlling \textbf{\textit{soil erosion}}, and as revegetator, often after \textbf{\textit{forest fires}}.} \\
&  B. a decrease in the amount of erosion. &  \\
&  C. an increase in the bacterium population. &  \\
&  D. an increase in the production of oxygen and fire. &  \\
\\ 
\midrule
\multirow{4}{1.0\linewidth}{Juan and LaKeisha roll a few objects down a ramp. They want to see which object rolls the farthest. What should they do so they can repeat their \textbf{\textit{investigation?}}} 
& A. Put the objects in groups. & \multirow{4}{1.0\linewidth}{The use of measurement developed to allow \textbf{\textit{recording}} and comparison of \textbf{\textit{observations}} made at different times and places, by different people.} \\
& B. Change the height of the ramp. &  \\
& C. Choose different objects to roll. &  \\
& D. \textbf{\textit{Record}} the details of the \textbf{\textit{investigation}}. $\checkmark$ &  \\
\\
\midrule
\multirow{4}{1.0\linewidth}{Which statement best explains why the sun appears to \textbf{\textit{move across the sky}} each day?}
& A. The sun revolves around Earth.
 & \multirow{4}{1.0\linewidth}{\textbf{\textit{Earth's rotation}} about its \textbf{\textit{axis}} causes the fixed stars to apparently \textbf{\textit{move across the sky}} in a way that depends on the observer's  latitude.} \\
& B. Earth rotates around the sun. &  \\
& C. The sun revolves on its axis. &  \\
& D. \textbf{\textit{Earth rotates}} on its \textbf{\textit{axis}}. $\checkmark$ &  \\
\bottomrule
\end{tabular}
\caption{Examples of corrected errors using the reference corpus enriched by the sentences from Wikipedia.}
\label{tab:eval:wiki_gain}
\end{table*}

We see further improvements on ARC-Challenge and OpenBookQA, by fine-tuning the pre-fine-tuned model on multiple target datasets (\ie, ARC-Easy, ARC-Challenge, and OpenBookQA). However, we do not see a similar gain on ARC-Easy by increasing the number of in-domain training instances. We will further discuss it in Section~\ref{sec:exp:md}.

Our best models (\ie, $\text{IRC} + \text{IEC}$ for ARC-Easy and $\text{IRC} + \text{IEC} + \text{MD}$ for ARC-Challenge and OpenBookQA) outperform the strong baseline BERT$^{\text{\RNum{2}}}$ introduced in Section~\ref{sec:method:basic} ($74.7\%$ \vs~$71.9\%$ on ARC-Easy, $53.7\%$ \vs~$44.1\%$ on ARC-Challenge, and $68.0\%$ \vs~$64.8\%$ on OpenBookQA), which already beats the previous state-of-the-art model RS$^{\text{\RNum{2}}}$. In the remaining sections, we analyze our models and discuss the impacts of external knowledge from various aspects.

\subsection{Impact of External Knowledge from an Open-Domain Resource}
\label{sec:exp:edl}

Table~\ref{tab:eval:wiki_gain} shows some examples of errors produced by IRC (Table~\ref{tab:eval:transfer}) that do not leverage external knowledge from open-domain resources. These errors can be corrected by enriching the reference corpus with external sentences extracted from Wikipedia ($\text{IRC} + \text{IEC}$ in Table~\ref{tab:eval:transfer}). %
In the first example, the correct answer option \emph{``Kool-Aid''} never appears in the original reference corpus. As a result, without external background knowledge, it is less likely to infer that \emph{``Kool-Aid''} refers to liquid (can boil) here.

In addition to performing information retrieval on the enriched reference \emph{corpus}, we investigate an alternative approach that uses concept identification and linking to directly enrich the reference \emph{document} for each (question, answer option) pair.
More specifically, we apply concept identification and linking to each (question, answer option) pair $(q, o_i)$ and extract sentences from Wikipedia based on the linked concepts. These extracted sentences are appended to the reference documents $d_i$ of $(q, o_i)$ directly. We still keep up to $K$ (\ie, $50$) sentences per document. We observe that this direct appending approach generally cannot outperform the reference corpus enrichment approach described in Section~\ref{sec:method:open-domain}.

\begin{table}[t!]
\centering
\small
\begin{tabular}{lcccc}
\toprule
\bf Task  & \bf Wiki &  \bf OBQA & \bf ARC & \bf Total\\
\midrule
ARC-E    & 20.8  & 0.4  & 78.7    & 1,039,059\\
ARC-C    & 21.5  & 0.4  & 78.2    & 517,846 \\
OBQA     & 20.6  & 1.1  & 78.3    & 1,191,347\\ 
\bottomrule
\end{tabular}
\caption{Percentage (\%) of retrieved sentences from each source. Wiki: Wikipedia; Total: total number of retrieved sentences for all (question, answer option) pairs in a single task. ARC-Easy and ARC-Challenge share the same original reference corpus.}
\label{tab:eval:percentage}
\end{table}

We report the statistics of the sentences (without redundancy removal) extracted from each source in Table~\ref{tab:eval:percentage}, used as inputs to our methods IRC + IEC and IRC + IEC + MD in Table~\ref{tab:eval:transfer}. As the original reference corpus of OpenBookQA is made up of 1,326 sentences, fewer retrieved sentences are extracted from its reference corpus for all tasks compared to other sources.

\subsection{Impact of External Knowledge from In-Domain Data}
\label{sec:exp:md}

\begin{table}[t!]
\centering
\small
\begin{tabular}{lccc}
\toprule
\bf First 4  & \bf Last 4 &  \bf Accuracy & \bf \# Epochs \\
\midrule
ARC-C    & ARC-E & 69.4  & 8 \\
OBQA    & ARC-E & 70.9  & 8 \\
ARC-C + OBQA   & ARC-E & 72.6  & 8 \\ 
\midrule
ARC-E  & - & 72.9 & 4 \\
ARC-E  & ARC-E & 74.7 & 8 \\
\bottomrule
\end{tabular}
\caption{Accuracy (\%) on the ARC-Easy test set. The first four epochs are fine-tuned using the dataset(s) in the first column. The last four epochs are fine-tuned using the dataset in the second column. \# Epochs: the total number of epochs.}
\label{tab:eval:4epochs}
\end{table}

Compared to fine-tuning the pre-fine-tuned model on a single multiple-choice subject-area QA dataset, we observe improvements in accuracy by fine-tuning on multiple in-domain datasets (MD) simultaneously (Section~\ref{sec:method:in-domain}) for ARC-Challenge and OpenBookQA. In particular, we see a dramatic gain on the ARC-Challenge dataset (from $46.1\%$ to $53.7\%$) as shown in Table~\ref{tab:eval:transfer}.

However, MD leads to a performance drop on ARC-Easy. We hypothesize that other commonly adopted approaches may also lead to performance drops. To verify that, we explore another way of utilizing external knowledge for ARC-Easy by first fine-tuning the pre-fine-tuned model for four epochs on external in-domain data (\ie, ARC-Challenge, OpenBookQA, or ARC-Challenge $+$ OpenBookQA) and then further fine-tuning for four epochs on ARC-Easy. As shown in Table~\ref{tab:eval:4epochs}, we also observe that compared to only fine-tuning on ARC-Easy, fine-turning on external in-domain data hurts the performance. The consistent performance drops across the two methods of using MD on ARC-Easy are perhaps due to an intrinsic property of the tasks themselves -- the question-answer instances in ARC-Easy are relatively simpler than those in ARC-Challenge and OpenBookQA. Introducing relatively complex problems from ARC-Challenge and OpenBookQA may hurt the final performance on ARC-Easy. As mentioned earlier, compared to questions in ARC-Easy, questions in ARC-Challenge are less likely to be answered correctly by retrieval-based or word co-occurrence methods. We argue that questions in the ARC-Challenge tend to require more external knowledge for reasoning, similar to the observation of~\newcite{sugawara2018makes} ($30.0\%$ vs. $20.0\%$).

\subsection{Discussions about Question Types and Remaining Challenges}

We use the human annotations such as required reasoning skills (\ie, \emph{word matching}, \emph{paraphrasing}, \emph{knowledge}, \emph{meta/whole}, and \emph{math/whole}) and validity of questions in ARC-Easy and ARC-Challenge released by~\newcite{sugawara2018makes} to analyze the impacts of external knowledge on instances in various categories. Sixty instances are annotated for each dataset. We refer readers to~\newcite{sugawara2018makes} for detailed definitions of each category. We do not report the accuracy for \emph{math/whole} as no annotated question in ARC belongs to this category.

We compare the BERT$^{\text{\RNum{2}}}$ baseline in Table~\ref{tab:eval:transfer} that only uses the original reference corpus of a given end task with our best model. As shown in Table~\ref{tab:eval:type}, by leveraging external knowledge from in-domain datasets (instances and reference corpora) and open-domain texts, we observe consistent improvements on most of the categories. Based on these experimental results on the annotated subset, we may assume it could be a promising direction to further improve challenging multiple-choice subject-area QA tasks through exploiting high-quality external knowledge besides designing task-specific models for different types of questions~\cite{clark2016combining}.

\begin{table}[t!]
\centering
\small
\begin{tabular}{lcccc}
\toprule
\multirow{2}{*}{\textbf{Question Type}} & \multicolumn{2}{c}{\textbf{ARC-E}} & \multicolumn{2}{c}{\textbf{ARC-C}} \\
                               & \bf BERT$^{\text{\RNum{2}}}$  & \bf Ours     & \bf BERT$^{\text{\RNum{2}}}$ & \bf Ours  \\
\midrule
Word Matching                  & 81.3         & \bf 85.4          & 30.4        & \bf 73.9                \\
Paraphrasing                   & 90.9         & 90.9              & 46.7        & \bf 66.7                \\
Knowledge                      & 58.3         & \bf 83.3          & 44.4        & \bf 55.6                \\
Math/Logic                     & 100.0        & 100.0             & 33.3        & 33.3                \\
\midrule
Valid                          & 80.0         & \bf 86.0          & 36.1        & \bf 66.7              \\
Invalid                        & 50.0         & \bf 80.0          & 41.7        & 41.7               \\
Easy                           & 80.0         & \bf 90.0          & 33.3        & \bf 53.3           \\
Hard                           & 70.0         & \bf 80.0          & 43.3        & \bf 60.0           \\
\bottomrule
\end{tabular}
\caption{Accuracy (\%) by different categories on the annotated test sets of ARC-Easy and ARC-Challenge, which are released by~\newcite{sugawara2018makes}.} 
\label{tab:eval:type}
\end{table}

We also analyze the instances that our approach fails to answer correctly in the OpenBookQA development set to study the remaining challenges. It might be promising to identify the relations among concepts within an answer option. For example, our current model mistakenly selects the answer option \emph{``the sun orbits the earth''} associated with the question \emph{``Revolution happens when ?''} probably because \emph{``sun''}, \emph{``orbits''}, and \emph{``earth''} frequently co-occur in our generated reference document, though these concepts such as \emph{``revolution''} are successfully linked to their corresponding Wikipedia pages in the astronomy field.

Besides, we might also need to identify causal relations between events. For example, given the question \emph{``The type of climate change known as anthropogenic is caused by this''}, our model mistakenly predicts another answer option \emph{``forest fires''} with its associated contexts \emph{``climate change has caused the island to suffer more frequent severe droughts, leading to large forest fires''}, instead of the real cause \emph{``humanity''} supported by \emph{``the problem now is with anthropogenic climate change---that is, climate change caused by human activity, which is making the climate change a lot faster than it normally would''}.

\begin{figure}[!t]
   \begin{center}
   \includegraphics[width=.9\linewidth]{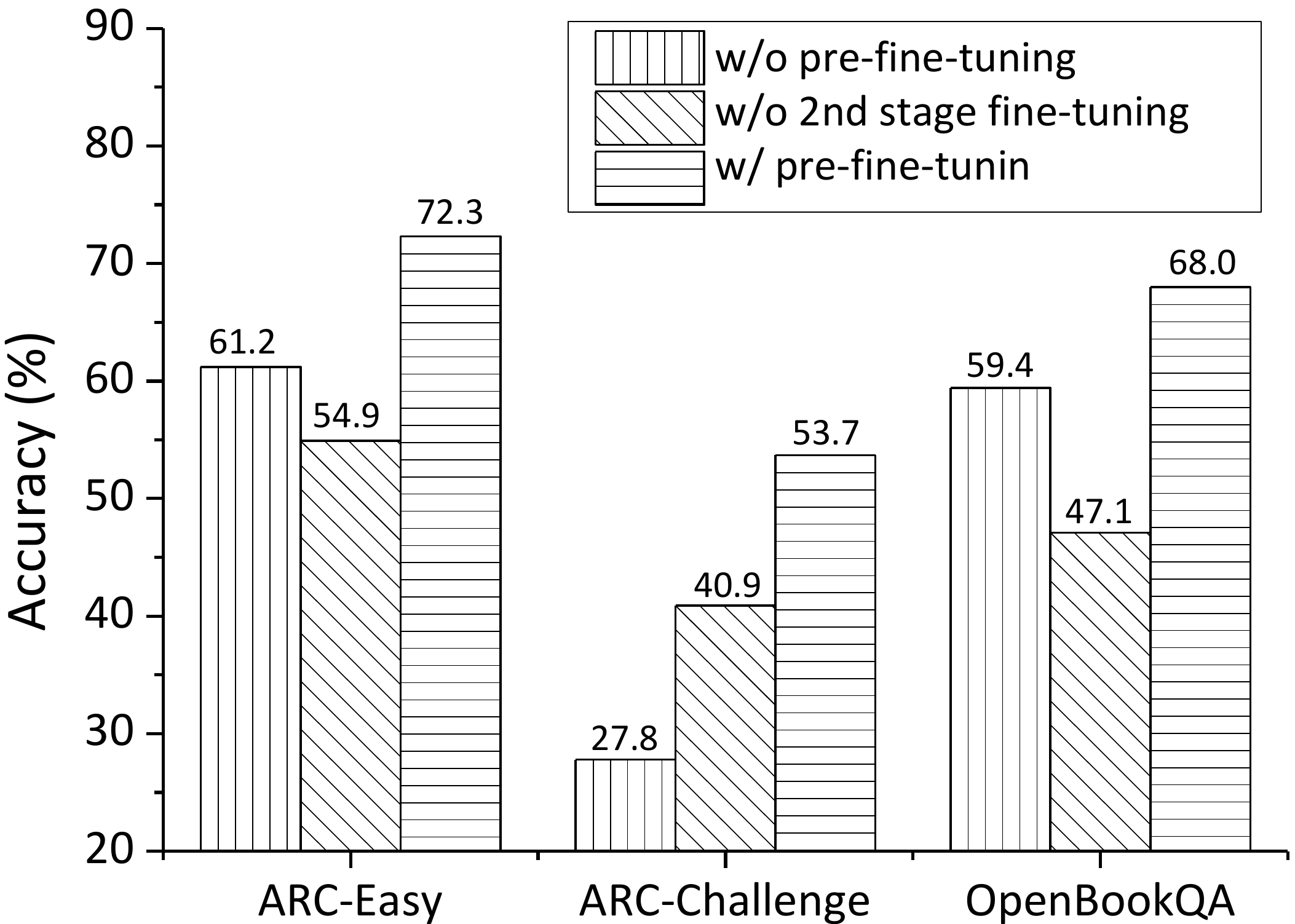}
   \end{center}
 \caption{Accuracy (\%) on the test sets of evaluation tasks with and without the pre-fine-tuning stage (2nd stage fine-tuning: fine-tune the pre-fine-tuned model on target science question answering datasets).}
 \label{fig:expt:pre-fine_tuning}
\end{figure}
\subsection{Discussions about Pre-Fine-Tuning}
\label{sec:exp:race}

Previous work~\cite{bert2018} has shown that fine-tuning $\text{BERT}_\text{LARGE}$ on small datasets can be sometimes unstable. Additionally,~\newcite{sun2018improving} show that fine-tuning GPT~\cite{radfordimproving} that is pre-fine-tuned on RACE can dramatically improve the performance of relatively small multiple-choice tasks. Here we only use the BERT$^{\text{\RNum{2}}}$ baseline for a brief discussion. We have a similar observation: we can obtain more stable performance on the target datasets by first fine-tuning BERT on RACE (language exams), and we see consistent performance improvements on all the evaluated science QA datasets. As shown in Figure~\ref{fig:expt:pre-fine_tuning}, we see that the performance drops dramatically without using pre-fine-tuning on the RACE dataset. 
\section{Related Work}

\subsection{Subject-Area QA Tasks and Methods}

As there is not a clear distinction between QA and machine reading comprehension (\textbf{MRC}) tasks, for convenience we call a task in which there is no reference document provided for each instance as a QA task. 
In this paper, we focus on multiple-choice subject-area QA tasks, where the in-domain reference corpus does not provide sufficient relevant content on its own to answer a significant portion of the questions~\cite{clark2016combining,kobayashi2017automated,welbl2017crowdsourcing,clark2018think,mihaylov2018can}. In contrast to other types of QA scenarios~\cite{nguyen2016ms,dhingra2017quasar,triviaQA,dunn2017searchqa,nqgoogle}, in this setting: (1) the reference corpus does not reliably contain text spans from which the answers can be drawn, and (2) it does not provide sufficient information on its own to answer a significant portion of the questions. Thus they are suitable for us to study how to exploit external knowledge for QA.

Our work follows the general framework of discriminatively fine-tuning a pre-trained language model such as GPT~\cite{radfordimproving} and BERT~\cite{bert2018} on QA tasks~\cite{radfordimproving,bert2018,hu2018read,yang2019end}. As shown in Table~\ref{tab:eval:transfer}, the baseline based on BERT already outperforms previous state-of-the-art methods designed for subject-area QA tasks~\cite{yadav2018sanity,pirtoaca2018improving,ni2018learning,sun2018improving}.

\subsection{Utilization of External Knowledge for Subject-Area QA}

Previous studies have explored many ways to leverage structured knowledge to solve questions in subject areas such as science exams. Many researchers investigate how to directly or indirectly use automatically constructed knowledge bases/graphs from reference corpora~\cite{khot2017answering,kwon2018controlling,khashabi2018question,zhang2018kg} or existing external general knowledge graphs~\cite{li2015answering,sachan2016science,wang2018yuanfudao,wang2018improving,zhong2018improving,musa2018answering} such as ConceptNet~\cite{speer2017conceptnet}. However, for subject-area QA, unstructured knowledge is seldom considered in previous studies, and it is still not clear the usefulness of this kind of knowledge.

As far as we know, for subject-area QA tasks, this is the first attempt to impart sources of external unstructured knowledge into one state-of-the-art pre-trained language model, and we are among the first to investigate the effectiveness of the external unstructured texts in Wikipedia~\cite{pirtoaca2018improving} and additional in-domain QA data.

\subsection{Utilization of External Knowledge for Other Types of QA and MRC}
For both QA and MRC tasks in which the majority of answers are extractive such as SQuAD~\cite{rajpurkar2016squad} and TriviaQA~\cite{triviaQA}, previous work has shown that it is useful to introduce external open-domain QA instances and textual information from Wikipedia by first retrieving relevant documents in Wikipedia and then running a MRC model to extract a text span from the documents based on the question~\cite{chen2017reading,wang2017r,kratzwald2018adaptive,lee2018ranking,lin2018denoising}.

Based on Wikipedia, we apply concept identification and linking to enrich QA reference corpora, which has not been explored before. Compared to previous data argumentation studies for other types of QA tasks~\cite{yu2018qanet}, differences exist in: 1) we focus on in-domain data and discuss the impacts of the difficulties of additional in-domain instances on a target task; 2) we are the first to show it is useful to merge reference corpora from different in-domain subject-area QA tasks.

\section{Conclusion and Future Work}

We focus on how to incorporate external knowledge into a pre-trained model to improve subject-area QA tasks that require background knowledge. We exploit two sources of external knowledge through: enriching the original reference corpus with relevant texts from open-domain Wikipedia and using additional in-domain QA datasets (instances and reference corpora) for training. Experimental results on ARC-Easy, ARC-Challenge, and OpenBookQA show the effectiveness of our simple method. The promising results also demonstrate the importance of unstructured external knowledge for subject-area QA. In the future, we plan to jointly exploit various types of external unstructured and structured knowledge. %

\section*{Acknowledgments}
We thank the anonymous reviewers for their constructive and helpful feedback.

\bibliography{emnlp2019}

\begin{thebibliography}{48}
\expandafter\ifx\csname natexlab\endcsname\relax\def\natexlab#1{#1}\fi

\bibitem[{Bird and Loper(2004)}]{bird2004nltk}
Steven Bird and Edward Loper. 2004.
\newblock \href {http://www.aclweb.org/anthology/P04-3031} {{NLTK}: the natural
  language toolkit}.
\newblock In \emph{Proceedings of the ACL (Demonstrations)}, Barcelona, Spain.

\bibitem[{Chen et~al.(2017)Chen, Fisch, Weston, and Bordes}]{chen2017reading}
Danqi Chen, Adam Fisch, Jason Weston, and Antoine Bordes. 2017.
\newblock \href {http://www.aclweb.org/anthology/P17-1171} {Reading {Wikipedia}
  to answer open-domain questions}.
\newblock In \emph{Proceedings of the ACL}, Vancouver, Canada.

\bibitem[{Clark et~al.(2018)Clark, Cowhey, Etzioni, Khot, Sabharwal, Schoenick,
  and Tafjord}]{clark2018think}
Peter Clark, Isaac Cowhey, Oren Etzioni, Tushar Khot, Ashish Sabharwal, Carissa
  Schoenick, and Oyvind Tafjord. 2018.
\newblock \href {https://arxiv.org/abs/1803.05457v1} {Think you have solved
  question answering? {Try ARC}, the {AI2} reasoning challenge}.
\newblock \emph{arXiv preprint}, cs.CL/1803.05457v1.

\bibitem[{Clark et~al.(2016)Clark, Etzioni, Khot, Sabharwal, Tafjord, Turney,
  and Khashabi}]{clark2016combining}
Peter Clark, Oren Etzioni, Tushar Khot, Ashish Sabharwal, Oyvind Tafjord,
  Peter~D Turney, and Daniel Khashabi. 2016.
\newblock \href
  {http://www.aaai.org/ocs/index.php/AAAI/AAAI16/paper/download/11963/11990}
  {Combining retrieval, statistics, and inference to answer elementary science
  questions.}
\newblock In \emph{Proceedings of the AAAI}, Phoenix, AZ.

\bibitem[{Devlin et~al.(2019)Devlin, Chang, Lee, and Toutanova}]{bert2018}
Jacob Devlin, Ming-Wei Chang, Kenton Lee, and Kristina Toutanova. 2019.
\newblock \href {https://arxiv.org/abs/1810.04805v1} {{BERT}: Pre-training of
  deep bidirectional transformers for language understanding}.
\newblock In \emph{Proeedings of the NAACL-HLT}, Minneapolis, MN.

\bibitem[{Dhingra et~al.(2017)Dhingra, Mazaitis, and Cohen}]{dhingra2017quasar}
Bhuwan Dhingra, Kathryn Mazaitis, and William~W Cohen. 2017.
\newblock \href {https://arxiv.org/abs/1707.03904v2} {Quasar: Datasets for
  question answering by search and reading}.
\newblock \emph{arXiv preprint}, cs.CL/1707.03904v2.

\bibitem[{Dunn et~al.(2017)Dunn, Sagun, Higgins, Guney, Cirik, and
  Cho}]{dunn2017searchqa}
Matthew Dunn, Levent Sagun, Mike Higgins, V~Ugur Guney, Volkan Cirik, and
  Kyunghyun Cho. 2017.
\newblock \href {https://arxiv.org/abs/1704.05179v3} {{SearchQA}: A new {Q\&A}
  dataset augmented with context from a search engine}.
\newblock \emph{arXiv preprint}, cs.CL/1704.05179v3.

\bibitem[{Hu et~al.(2019)Hu, Peng, Huang, Yang, Zhou et~al.}]{hu2018read}
Minghao Hu, Yuxing Peng, Zhen Huang, Nan Yang, Ming Zhou, et~al. 2019.
\newblock \href {https://arxiv.org/abs/1808.05759v5} {Read+{V}erify: Machine
  reading comprehension with unanswerable questions}.
\newblock In \emph{Proceedings of the AAAI}, Honolulu, HI.

\bibitem[{Joshi et~al.(2017)Joshi, Choi, Weld, and Zettlemoyer}]{triviaQA}
Mandar Joshi, Eunsol Choi, Daniel~S. Weld, and Luke Zettlemoyer. 2017.
\newblock \href {http://arxiv.org/abs/1705.03551v2} {Trivia{QA}: {A} large
  scale distantly supervised challenge dataset for reading comprehension}.
\newblock \emph{arXiv preprint}, cs.CL/1705.03551v2.

\bibitem[{Kendeou and Van Den~Broek(2007)}]{kendeou2007effects}
Panayiota Kendeou and Paul Van Den~Broek. 2007.
\newblock \href {https://link.springer.com/article/10.3758/BF03193491} {The
  effects of prior knowledge and text structure on comprehension processes
  during reading of scientific texts}.
\newblock \emph{Memory \& cognition}, 35(7).

\bibitem[{Khashabi et~al.(2017)Khashabi, Khot, Sabharwal, and
  Roth}]{khashabi2017learning}
Daniel Khashabi, Tushar Khot, Ashish Sabharwal, and Dan Roth. 2017.
\newblock \href {http://www.aclweb.org/anthology/K17-1010} {Learning what is
  essential in questions}.
\newblock In \emph{Proceedings of the CoNLL 2017}.

\bibitem[{Khashabi et~al.(2018)Khashabi, Khot, Sabharwal, and
  Roth}]{khashabi2018question}
Daniel Khashabi, Tushar Khot, Ashish Sabharwal, and Dan Roth. 2018.
\newblock \href
  {http://www.cis.upenn.edu/~danielkh/files/2018_semanticilp/2018_aaai_semanticilp.pdf}
  {Question answering as global reasoning over semantic abstractions}.
\newblock In \emph{Proceedings of the AAAI}, New Orleans, LA.

\bibitem[{Khot et~al.(2017)Khot, Sabharwal, and Clark}]{khot2017answering}
Tushar Khot, Ashish Sabharwal, and Peter Clark. 2017.
\newblock \href {http://www.aclweb.org/anthology/P17-2049} {Answering complex
  questions using open information extraction}.
\newblock In \emph{Proceedings of the ACL}, Vancouver, Canada.

\bibitem[{Khot et~al.(2018)Khot, Sabharwal, and Clark}]{khot2018scitail}
Tushar Khot, Ashish Sabharwal, and Peter Clark. 2018.
\newblock \href
  {https://www.aaai.org/ocs/index.php/AAAI/AAAI18/paper/viewPaper/17368}
  {{SciTail}: A textual entailment dataset from science question answering}.
\newblock In \emph{Proceedings of the AAAI}, New Orleans, LA.

\bibitem[{Kobayashi et~al.(2017)Kobayashi, Ishii, Hoshino, Miyashita, and
  Matsuzaki}]{kobayashi2017automated}
Mio Kobayashi, Ai~Ishii, Chikara Hoshino, Hiroshi Miyashita, and Takuya
  Matsuzaki. 2017.
\newblock \href {http://www.aclweb.org/anthology/I17-1097} {Automated
  historical fact-checking by passage retrieval, word statistics, and virtual
  question-answering}.
\newblock In \emph{Proceedings of the IJCNLP}, Taipei, Taiwan.

\bibitem[{Kratzwald and Feuerriegel(2018)}]{kratzwald2018adaptive}
Bernhard Kratzwald and Stefan Feuerriegel. 2018.
\newblock \href {http://www.aclweb.org/anthology/D18-1055} {Adaptive document
  retrieval for deep question answering}.
\newblock In \emph{Proceedings of the EMNLP}, Brussels, Belgium.

\bibitem[{Kwiatkowski et~al.(2019)Kwiatkowski, Palomaki, Redfield, Collins,
  Parikh, Alberti, Epstein, Polosukhin, Kelcey, Devlin, Lee, Toutanova, Jones,
  Chang, Dai, Uszkoreit, Le, and Petrov}]{nqgoogle}
Tom Kwiatkowski, Jennimaria Palomaki, Olivia Redfield, Michael Collins, Ankur
  Parikh, Chris Alberti, Danielle Epstein, Illia Polosukhin, Matthew Kelcey,
  Jacob Devlin, Kenton Lee, Kristina~N. Toutanova, Llion Jones, Ming-Wei Chang,
  Andrew Dai, Jakob Uszkoreit, Quoc Le, and Slav Petrov. 2019.
\newblock \href
  {https://storage.googleapis.com/pub-tools-public-publication-data/pdf/b8c26e4347adc3453c15d96a09e6f7f102293f71.pdf}
  {{Natural Questions}: A benchmark for question answering research}.
\newblock \emph{TACL}.

\bibitem[{Kwon et~al.(2018)Kwon, Trivedi, Jansen, Surdeanu, and
  Balasubramanian}]{kwon2018controlling}
Heeyoung Kwon, Harsh Trivedi, Peter Jansen, Mihai Surdeanu, and Niranjan
  Balasubramanian. 2018.
\newblock \href
  {https://link.springer.com/chapter/10.1007/978-3-319-76941-7_72} {Controlling
  information aggregation for complex question answering}.
\newblock In \emph{Proceedings of the ECIR}, Grenoble, France.

\bibitem[{Lai et~al.(2017)Lai, Xie, Liu, Yang, and Hovy}]{lai2017race}
Guokun Lai, Qizhe Xie, Hanxiao Liu, Yiming Yang, and Eduard Hovy. 2017.
\newblock \href {http://www.aclweb.org/anthology/D17-1082} {{RACE}: Large-scale
  reading comprehension dataset from examinations}.
\newblock In \emph{Proceedings of the EMNLP}, Copenhagen, Denmark.

\bibitem[{Lee et~al.(2018)Lee, Yun, Kim, Ko, and Kang}]{lee2018ranking}
Jinhyuk Lee, Seongjun Yun, Hyunjae Kim, Miyoung Ko, and Jaewoo Kang. 2018.
\newblock \href {http://www.aclweb.org/anthology/D18-1053} {Ranking paragraphs
  for improving answer recall in open-domain question answering}.
\newblock In \emph{Proceedings of the EMNLP}, Brussels, Belgium.

\bibitem[{Li and Clark(2015)}]{li2015answering}
Yang Li and Peter Clark. 2015.
\newblock \href {http://www.aclweb.org/anthology/D15-1236} {Answering
  elementary science questions by constructing coherent scenes using background
  knowledge}.
\newblock In \emph{Proceedings of the EMNLP}, Lisbon, Portugal.

\bibitem[{Lin et~al.(2018)Lin, Ji, Liu, and Sun}]{lin2018denoising}
Yankai Lin, Haozhe Ji, Zhiyuan Liu, and Maosong Sun. 2018.
\newblock \href {http://www.aclweb.org/anthology/P18-1161} {Denoising distantly
  supervised open-domain question answering}.
\newblock In \emph{Proceedings of the ACL}, Melbourne, Australia.

\bibitem[{Manning et~al.(2014)Manning, Surdeanu, Bauer, Finkel, Bethard, and
  McClosky}]{corenlp}
Christopher~D. Manning, Mihai Surdeanu, John Bauer, Jenny Finkel, Steven~J.
  Bethard, and David McClosky. 2014.
\newblock \href {http://www.aclweb.org/anthology/P/P14/P14-5010} {The
  {Stanford} {CoreNLP} natural language processing toolkit}.
\newblock In \emph{Proceedings of the ACL (Demonstrations)}, Baltimore, MD.

\bibitem[{McCandless et~al.(2010)McCandless, Hatcher, and Gospodnetic}]{lucene}
Michael McCandless, Erik Hatcher, and Otis Gospodnetic. 2010.
\newblock \href {https://dl.acm.org/citation.cfm?id=1893016} {\emph{Lucene in
  Action, Second Edition: Covers Apache Lucene 3.0}}.
\newblock Manning Publications Co., Greenwich, CT.

\bibitem[{McNamara et~al.(2004)McNamara, Levinstein, and
  Boonthum}]{mcnamara2004istart}
Danielle~S McNamara, Irwin~B Levinstein, and Chutima Boonthum. 2004.
\newblock \href {https://link.springer.com/article/10.3758/BF03195567}
  {{iSTART}: Interactive strategy training for active reading and thinking}.
\newblock \emph{Behavior Research Methods, Instruments, \& Computers}, 36(2).

\bibitem[{Mihaylov et~al.(2018)Mihaylov, Clark, Khot, and
  Sabharwal}]{mihaylov2018can}
Todor Mihaylov, Peter Clark, Tushar Khot, and Ashish Sabharwal. 2018.
\newblock \href {http://aclweb.org/anthology/D18-1260} {Can a suit of armor
  conduct electricity? a new dataset for open book question answering}.
\newblock In \emph{Proceedings of the EMNLP}, Brussels, Belgium.

\bibitem[{Musa et~al.(2018)Musa, Wang, Fokoue, Mattei, Chang, Kapanipathi,
  Makni, Talamadupula, and Witbrock}]{musa2018answering}
Ryan Musa, Xiaoyan Wang, Achille Fokoue, Nicholas Mattei, Maria Chang, Pavan
  Kapanipathi, Bassem Makni, Kartik Talamadupula, and Michael Witbrock. 2018.
\newblock \href {https://arxiv.org/abs/1809.05726v1} {Answering science exam
  questions using query rewriting with background knowledge}.
\newblock \emph{arXiv preprint}, cs.AI/1809.05726v1.

\bibitem[{Nguyen et~al.(2016)Nguyen, Rosenberg, Song, Gao, Tiwary, Majumder,
  and Deng}]{nguyen2016ms}
Tri Nguyen, Mir Rosenberg, Xia Song, Jianfeng Gao, Saurabh Tiwary, Rangan
  Majumder, and Li~Deng. 2016.
\newblock \href {https://arxiv.org/abs/1611.09268v2} {{MS MARCO}: A human
  generated machine reading comprehension dataset}.
\newblock \emph{arXiv preprint}, cs.CL/1611.09268v2.

\bibitem[{Ni et~al.(2019)Ni, Zhu, Chen, and McAuley}]{ni2018learning}
Jianmo Ni, Chenguang Zhu, Weizhu Chen, and Julian McAuley. 2019.
\newblock \href {https://arxiv.org/abs/1808.09492v4} {Learning to attend on
  essential terms: An enhanced retriever-reader model for open-domain question
  answering}.
\newblock In \emph{Proceedings of the NAACL-HLT}, Minneapolis, MN.

\bibitem[{Pan et~al.(2015)Pan, Cassidy, Hermjakob, Ji, and Knight}]{Pan2015}
Xiaoman Pan, Taylor Cassidy, Ulf Hermjakob, Heng Ji, and Kevin Knight. 2015.
\newblock \href {http://www.aclweb.org/anthology/N15-1119} {Unsupervised entity
  linking with abstract meaning representation}.
\newblock In \emph{Proceedings of the NAACL-HLT}, Denver, CO.

\bibitem[{Pirtoaca et~al.(2019)Pirtoaca, Rebedea, and
  Ruseti}]{pirtoaca2018improving}
George-Sebastian Pirtoaca, Traian Rebedea, and Stefan Ruseti. 2019.
\newblock \href {https://arxiv.org/abs/1812.02971v2} {Improving retrieval-based
  question answering with deep inference models}.
\newblock \emph{arXiv preprint}, cs.CL/1812.02971v2.

\bibitem[{Radford et~al.(2018)Radford, Narasimhan, Salimans, and
  Sutskever}]{radfordimproving}
Alec Radford, Karthik Narasimhan, Tim Salimans, and Ilya Sutskever. 2018.
\newblock \href
  {https://s3-us-west-2.amazonaws.com/openai-assets/research-covers/language-unsupervised/language_understanding_paper.pdf}
  {Improving language understanding by generative pre-training}.
\newblock In \emph{Preprint}.

\bibitem[{Rajpurkar et~al.(2016)Rajpurkar, Zhang, Lopyrev, and
  Liang}]{rajpurkar2016squad}
Pranav Rajpurkar, Jian Zhang, Konstantin Lopyrev, and Percy Liang. 2016.
\newblock \href {http://www.aclweb.org/anthology/D16-1264} {{SQuAD}: 100,000+
  questions for machine comprehension of text}.
\newblock In \emph{Proceedings of the EMNLP}, Austin, TX.

\bibitem[{Sachan et~al.(2016)Sachan, Dubey, and Xing}]{sachan2016science}
Mrinmaya Sachan, Avinava Dubey, and Eric~P Xing. 2016.
\newblock \href {http://www.aclweb.org/anthology/P16-2076} {Science question
  answering using instructional materials}.
\newblock In \emph{Proceedings of the ACL}, Berlin, Germany.

\bibitem[{Salmer{\'o}n et~al.(2006)Salmer{\'o}n, Kintsch, and
  Ca{\~a}s}]{salmeron2006reading}
Ladislao Salmer{\'o}n, Walter Kintsch, and Jos{\'e}~J Ca{\~a}s. 2006.
\newblock \href {https://link.springer.com/article/10.3758/BF03193262} {Reading
  strategies and prior knowledge in learning from hypertext}.
\newblock \emph{Memory \& Cognition}, 34(5).

\bibitem[{Speer et~al.(2017)Speer, Chin, and Havasi}]{speer2017conceptnet}
Robyn Speer, Joshua Chin, and Catherine Havasi. 2017.
\newblock \href
  {https://www.aaai.org/ocs/index.php/AAAI/AAAI17/paper/viewPDFInterstitial/14972/14051}
  {{ConceptNet} 5.5: An open multilingual graph of general knowledge}.
\newblock In \emph{Proceedings of the AAAI}, San Francisco, CA.

\bibitem[{Sugawara et~al.(2018)Sugawara, Inui, Sekine, and
  Aizawa}]{sugawara2018makes}
Saku Sugawara, Kentaro Inui, Satoshi Sekine, and Akiko Aizawa. 2018.
\newblock \href {http://aclweb.org/anthology/D18-1453} {What makes reading
  comprehension questions easier?}
\newblock In \emph{Proceedings of the EMNLP}, Brussels, Belgium.

\bibitem[{Sun et~al.(2019)Sun, Yu, Yu, and Cardie}]{sun2018improving}
Kai Sun, Dian Yu, Dong Yu, and Claire Cardie. 2019.
\newblock \href {https://arxiv.org/abs/1810.13441v2} {Improving machine reading
  comprehension with general reading strategies}.
\newblock In \emph{Proceedings of the NAACL-HLT}, Minneapolis, MN.

\bibitem[{Wang et~al.(2018{\natexlab{a}})Wang, Sun, Zhao, Shen, and
  Liu}]{wang2018yuanfudao}
Liang Wang, Meng Sun, Wei Zhao, Kewei Shen, and Jingming Liu.
  2018{\natexlab{a}}.
\newblock \href {http://aclweb.org/anthology/S18-1120} {Yuanfudao at
  {SemEval-2018 Task 11}: Three-way attention and relational knowledge for
  commonsense machine comprehension}.
\newblock In \emph{Proceedings of the SemEval}, New Orleans, LA.

\bibitem[{Wang et~al.(2018{\natexlab{b}})Wang, Yu, Guo, Wang, Klinger, Zhang,
  Chang, Tesauro, Zhou, and Jiang}]{wang2017r}
Shuohang Wang, Mo~Yu, Xiaoxiao Guo, Zhiguo Wang, Tim Klinger, Wei Zhang, Shiyu
  Chang, Gerald Tesauro, Bowen Zhou, and Jing Jiang. 2018{\natexlab{b}}.
\newblock \href {https://arxiv.org/abs/1709.00023v2} {${R}^{3}$: Reinforced
  reader-ranker for open-domain question answering}.
\newblock In \emph{Proceedings of the AAAI}, New Orleans, LA.

\bibitem[{Wang et~al.(2018{\natexlab{c}})Wang, Kapanipathi, Musa, Yu,
  Talamadupula, Abdelaziz, Chang, Fokoue, Makni, Mattei
  et~al.}]{wang2018improving}
Xiaoyan Wang, Pavan Kapanipathi, Ryan Musa, Mo~Yu, Kartik Talamadupula, Ibrahim
  Abdelaziz, Maria Chang, Achille Fokoue, Bassem Makni, Nicholas Mattei, et~al.
  2018{\natexlab{c}}.
\newblock \href {https://arxiv.org/abs/1809.05724v2} {Improving natural
  language inference using external knowledge in the science questions domain}.
\newblock \emph{arXiv preprint}, cs.CL/1809.05724v2.

\bibitem[{Welbl et~al.(2017)Welbl, Liu, and Gardner}]{welbl2017crowdsourcing}
Johannes Welbl, Nelson~F Liu, and Matt Gardner. 2017.
\newblock \href {http://www.aclweb.org/anthology/W17-4413} {Crowdsourcing
  multiple choice science questions}.
\newblock In \emph{Proceedings of the W-NUT}, Copenhagen, Denmark.

\bibitem[{Yadav et~al.(2019)Yadav, Bethard, and Surdeanu}]{yadavalignment}
Vikas Yadav, Steven Bethard, and Mihai Surdeanu. 2019.
\newblock \href {https://www.aclweb.org/anthology/N19-1274} {Alignment over
  heterogeneous embeddings for question answering}.
\newblock In \emph{Proceedings of the NAACL-HLT}, Minneapolis, MN.

\bibitem[{Yadav et~al.(2018)Yadav, Sharp, and Surdeanu}]{yadav2018sanity}
Vikas Yadav, Rebecca Sharp, and Mihai Surdeanu. 2018.
\newblock \href {https://dl.acm.org/citation.cfm?id=3210142} {Sanity check: A
  strong alignment and information retrieval baseline for question answering}.
\newblock In \emph{Proceedings of the ACM SIGIR}, Ann Arbor, MI.

\bibitem[{Yang et~al.(2019)Yang, Xie, Lin, Li, Tan, Xiong, Li, and
  Lin}]{yang2019end}
Wei Yang, Yuqing Xie, Aileen Lin, Xingyu Li, Luchen Tan, Kun Xiong, Ming Li,
  and Jimmy Lin. 2019.
\newblock \href {https://arxiv.org/abs/1902.01718v1} {End-to-end open-domain
  question answering with bertserini}.
\newblock \emph{arXiv preprint}, cs.CL/1902.01718v1.

\bibitem[{Yu et~al.(2018)Yu, Dohan, Luong, Zhao, Chen, Norouzi, and
  Le}]{yu2018qanet}
Adams~Wei Yu, David Dohan, Minh-Thang Luong, Rui Zhao, Kai Chen, Mohammad
  Norouzi, and Quoc~V Le. 2018.
\newblock \href {https://arxiv.org/abs/1804.09541v1} {{QANet}: Combining local
  convolution with global self-attention for reading comprehension}.
\newblock In \emph{Proceedings of the ICLR}, Vancouver, Canada.

\bibitem[{Zhang et~al.(2018)Zhang, Dai, Toraman, and Song}]{zhang2018kg}
Yuyu Zhang, Hanjun Dai, Kamil Toraman, and Le~Song. 2018.
\newblock \href {https://arxiv.org/abs/1805.12393v1} {{KG}\^{} 2: Learning to
  reason science exam questions with contextual knowledge graph embeddings}.
\newblock \emph{arXiv preprint}, cs.LG/1805.12393v1.

\bibitem[{Zhong et~al.(2018)Zhong, Tang, Duan, Zhou, Wang, and
  Yin}]{zhong2018improving}
Wanjun Zhong, Duyu Tang, Nan Duan, Ming Zhou, Jiahai Wang, and Jian Yin. 2018.
\newblock \href {https://arxiv.org/abs/1809.03568v1} {Improving question
  answering by commonsense-based pre-training}.
\newblock \emph{arXiv preprint}, cs.CL/1809.03568v1.

\end{thebibliography}
\bibliographystyle{acl_natbib}

\end{document}